\documentclass{article}

\PassOptionsToPackage{numbers, compress}{natbib}

\usepackage[final]{neurips_2024}

\usepackage{amsmath} 
\usepackage[utf8]{inputenc} 
\usepackage[T1]{fontenc}    
\usepackage{url}            
\usepackage{booktabs}       
\usepackage{amsfonts}       
\usepackage{nicefrac}       
\usepackage{microtype}      
\usepackage[numbers]{natbib}
\usepackage[dvipsnames,table,xcdraw]{xcolor}
\definecolor{cvprblue}{rgb}{0.21,0.49,0.74}
\usepackage{graphicx}
\usepackage{caption}
\usepackage{arydshln}
\usepackage{tikz}
\usepackage{wrapfig}
\usepackage{float}
\usepackage{multirow}
\usepackage{pifont}
\usepackage{makecell}
\usepackage{subfigure}

\usepackage{xcolor}

\usepackage{tikz}
\usepackage{xcolor}

\usepackage{listings}

\usepackage{ulem}

\usepackage[pagebackref,breaklinks,colorlinks,citecolor=cvprblue]{hyperref}

\usepackage{mdframed}
\definecolor{mygray}{gray}{0.97}
\colorlet{shadecolor}{mygray}
\newmdenv[%
  backgroundcolor=mygray, 
  linewidth=0pt
]{newshaded}

\title{Skywork R1V2: Multimodal Hybrid Reinforcement Learning for Reasoning }

\author{
    \textnormal{Chris}\thanks{Equal contribution}, \quad \textnormal{Yichen Wei}\footnotemark[1], \quad \textnormal{Yi Peng}, \quad \textnormal{Xiaokun Wang}, \quad \textnormal{Weijie Qiu}, \quad \textnormal{Wei Shen}, \\
    \textnormal{Tianyidan Xie}, \quad \textnormal{Jiangbo Pei}, \quad \textnormal{Jianhao Zhang}, \quad \textnormal{Yunzhuo Hao}, \quad \textnormal{Xuchen Song}\thanks{Corresponding author}, \\
    \textnormal{Yang Liu}\footnotemark[2], \quad \textnormal{Yahui Zhou} \\
    \quad\\
    \textnormal{Skywork AI, Kunlun Inc.}\\
    {\textnormal{chris@kunlun-inc.com, xuchen.song@kunlun-inc.com}}
}

\begin{document}

\maketitle

\begin{abstract}

We present Skywork R1V2, a next-generation multimodal reasoning model and a major leap forward from its predecessor, Skywork R1V. At its core, R1V2 introduces a hybrid reinforcement learning paradigm that jointly leverages the Mixed Preference Optimization (MPO) and the Group Relative Policy Optimization (GRPO), which harmonizes reward-model guidance with rule-based strategies, thereby addressing the long-standing challenge of balancing sophisticated reasoning capabilities with broad generalization. To further enhance training efficiency, we propose the Selective Sample Buffer (SSB) mechanism, which effectively addresses the vanishing advantages dilemma inherent in GRPO by prioritizing high-value samples throughout the optimization process. Notably, we observe that excessive reinforcement signals can induce visual hallucinations--a phenomenon we systematically monitor and mitigate through calibrated reward thresholds throughout the training process. Empirical results affirm the exceptional capability of R1V2, with benchmark-leading performances such as 62.6 on OlympiadBench, 78.9 on AIME2024, 63.6 on LiveCodeBench, and 73.6 on MMMU. These results underscore R1V2's superiority over existing open-source models and demonstrate significant progress in closing the performance gap with premier proprietary systems, including Gemini 2.5 and OpenAI-o4-mini. The Skywork R1V2 model weights have been publicly released to promote openness and reproducibility\footnote{\url{https://huggingface.co/Skywork/Skywork-R1V2-38B}}.

\end{abstract}

\section{Introduction}
Recent advancements in ``slow-thinking'' multimodal models—such as OpenAI-o1 \cite{openai2024gpt4o}, Gemini-Thinking \cite{google2024gemini2}, Kimi-1.5 \cite{team2025kimi}, and Skywork-R1V \cite{peng2025skywork}—have led to substantial progress in complex reasoning tasks across mathematics and science. These models emulate reflective cognitive processes, achieving stronger performance than their ``fast-thinking'' counterparts (e.g., GPT-4o \cite{openai2024gpt4o} and Claude-3.5 \cite{Claude2024}), which prioritize rapid generation over deliberate reasoning. For instance, slow-thinking models demonstrate over 30\% improvement on benchmarks like AIME24 and AMC23, along with approximately 10\% gains on science-related benchmarks such as GPQA \cite{openai2024gpt4o,jaech2024openai,openai2025gpto4}.

However, extending slow-thinking strategies to multimodal domains introduces new challenges. While improvements are observed on visual reasoning tasks such as MMMU \cite{yue2024mmmu} and MathVision \cite{wang2024measuring}, performance degrades on general perception benchmarks (e.g., AI2D \cite{kembhavi2016diagram}). This decline is often accompanied by a rise in visual hallucinations, highlighting a critical issue: \textit{How can we effectively promote slow-thinking behavior in vision-language models (VLMs) without compromising their generalization capabilities?}

To tackle this problem, we introduce \textbf{Skywork R1V2}, a next-generation vision-language model that acquires multimodal reasoning skills directly via reinforcement learning (RL), bypassing the need for teacher model distillation. R1V2 is trained on a comprehensive dataset spanning visual perception, scientific inquiry, and abstract reasoning, encompassing both general-purpose and reasoning-intensive tasks. The training process utilizes a hybrid reinforcement learning paradigm that jointly leverages the \textit{Mixed Preference Optimization} (MPO) \cite{wang2025enhancingreasoningabilitymultimodal} and the \textit{Group Relative Policy Optimization} (GRPO) \cite{shao2024deepseekmathpushinglimitsmathematical}.

R1V2 first implements MPO with three core learning objectives: (1) the relative preferences between response pairs, (2) the absolute quality of individual responses, and (3) the process for generating preferred responses \cite{wang2025enhancingreasoningabilitymultimodal}. The preference signals is provided by our Skywork-VL Reward \cite{wang2025skywork}—a reward model capable of evaluating both multimodal understanding and reasoning tasks. Given these high-quality preference signals, MPO effectively mitigates hallucinations and overthinking in the generated outputs, leading to improved performance on general vision tasks.


To further boost the reasoning capabilities, we incorporate the GRPO algorithm, which computes
relative advantages by comparing candidate responses within the same query group. However, as
training progresses, candidate responses tend to converge (i.e., become uniformly correct or incorrect),
resulting in vanishing advantage signals and limited reasoning diversity. Building upon insights from recent work on this challenge \cite{wang2025vl}, we employ
the \textit{Selective Sample Buffer} (SSB), which caches high-quality training examples with non-zero
advantages and reintroduces them during policy updates. This mechanism increases gradient density,
enhances training stability, and encourages deeper reasoning paths. The hybrid of MPO and GRPO
enables a more effective balance between reasoning specialization and generalization.

Extensive evaluations demonstrate the effectiveness of R1V2 across multiple authoritative multimodal benchmarks, including 62.6\% on OlympiadBench \cite{he2024olympiadbench}, 78.9\% on AIME2024, 63.6\% on LiveCodeBench \cite{jain2024livecodebench}, and 73.6\% on MMMU. These results not only establish new open-source baselines but also substantially reduce the performance gap with proprietary state-of-the-art models such as Gemini 2.5 \cite{gemini2.5} and OpenAI-o4-mini \cite{openai2025gpto4}.

In summary, Skywork R1V2 offers a promising and scalable framework for training robust and reflective multimodal reasoning agents via direct reinforcement learning. It highlights the potential of unifying high-level reasoning and perceptual understanding, paving the way for the next generation of general-purpose AI systems.

\section{Related Work}

\subsection{Multimodal Reasoning Models}

Recent advances in multimodal AI have increasingly focused on enhancing reasoning capabilities across different modalities. Proprietary models like  Claude-3 \cite{Claude2024}, GPT-4o \cite{openai2024gpt4o}, and Gemini \cite{team2024gemini} have demonstrated remarkable capabilities, while the open-source community has responded with competitive alternatives such as LLaVA \cite{liu2023visual}, Qwen-VL \cite{Qwen-VL}, and InternVL \cite{chen2023internvl}.

Recent innovations have shifted toward slow-thinking approaches (e.g., OpenAI-o1 \cite{openai2024gpt4o}, Gemini-Thinking \cite{google2024gemini2}, and Kimi-1.5 \cite{team2025kimi}), which introduce specialized mechanisms for extended deliberation. These models allocate additional computational resources to complex reasoning tasks, demonstrating substantial improvements on mathematical and scientific benchmarks. The first generation of Skywork-R1V \cite{peng2025skywork} pioneered the direct application of text reasoning capabilities to vision through advanced adapter techniques, establishing a new paradigm for multimodal reasoning. Concurrently, VL-Rethinker \cite{wang2025vl} enhanced multimodal slow-thinking through reinforcement learning with Selective Sample Replay, achieving strong performance on mathematical reasoning benchmarks.

However, a common challenge across these approaches is maintaining balance between specialized reasoning abilities and general-purpose multimodal understanding. Models optimized heavily for mathematical reasoning often demonstrate degraded performance on everyday visual tasks, while general-purpose models struggle with complex analytical reasoning. Hence, addressing the trade-off between reasoning specialization and generalization is one of the key motivations behind our work.

\subsection{Preference Optimization in Multimodal Models}

Preference optimization has emerged as a powerful paradigm for aligning AI systems with human expectations, though its extension from language-only to multimodal reasoning presents significant challenges. While foundational techniques like RLHF \cite{christiano2017deep} and DPO \cite{rafailov2023direct} have transformed text-based models, their direct application to multimodal contexts requires substantial adaptations to accommodate the complexity of cross-modal interactions.
Recent works on multimodal preference learning \cite{sun2023aligning,wang2025enhancingreasoningabilitymultimodal} have incorporated human preferences into vision-language alignment, enhancing response quality in general multimodal tasks.
Despite these advances in addressing straightforward visual question-answering scenarios, the application of preference optimization to complex multimodal reasoning remains relatively underexplored, with two critical limitations hindering progress in this domain. First, the binary nature of typical preference pairs fails to capture the nuanced progression of complex reasoning paths, where multiple equally valid solutions may exist with different intermediate steps. Second, existing reward models predominantly evaluate textual quality in isolation, overlooking the crucial relationship between visual interpretation and logical inference that defines successful multimodal reasoning.

\section{Methodology}
This section introduces the core methodology behind Skywork R1V2. Building upon the original R1V series, R1V2 incorporates a hybrid training strategy combining reinforcement learning and reward-model-guided preference learning to better balance reasoning capability and general-purpose performance. 

\subsection{Efficient Multimodal Transfer via Modular Reassembly}

To reduce reliance on large-scale multimodal reasoning data, we decouple the alignment of visual-language representations from the preservation of reasoning capabilities. 
Specifically, we introduce a lightweight multi-layer perceptron (MLP) adapter, denote as $f_c$, to bridge a frozen vision encoder $f_v$ with a reasoning-capable language model $f_l$. Here we choose InternViT-6B~\cite{chen2023internvl} to be vision encoder and QwQ-32B~\cite{qwq32b} to be our language model. Formally, given a visual input $x_v$ and a text input $x_t$, the overall process represented as:
\begin{equation}
     y = f_l(f_c(f_v(x_v)), x_t) 
\end{equation}
where $f_v$ extracts visual features, $f_c$ adapts these features for capability with the language model, and $f_l$ integrates both the adapted visual features and the textual input $x_t$ to perform reasoning and generate the output $y$.


Unlike the first generation of R1V, R1V2 eliminates the supervised fine-tuning (SFT) stage. 
Recent finding suggests that SFT can inadvertently undermine the performance of subsequent reinforcement learning or reasoning processes~\cite{SFTorRL}, which can hinder the model's ability to develop genuine reasoning behaviors.
Rather than relying on SFT, R1V2 adopts a modular approach that directly connects a pretrained reasoning language model with a visual adapter. While this approach leads to a slight reduction in general visual understanding, it preserves the inherent reasoning ability of the language model and significantly benefits overall reasoning performance by avoiding the degradation introduced by SFT.


We systematically experimented with freezing and activating different model components, and observed a remarkable phenomenon: capabilities in text and vision exhibit high transferability—improvements in one modality directly benefit the other. Notably, while training the vision encoder alone yields limited gains, both adapter-only training and joint LLM+adapter training prove highly effective, suggesting that cross-modal alignment rather than visual encoding represents the critical bottleneck in multimodal reasoning.

\subsection{Mixed Preference Optimization} 

To achieve a strong model before reinforcement learning and balance reasoning with generalization, we employ Mixed Preference Optimization (MPO) \cite{wang2025enhancingreasoningabilitymultimodal}, which demonstrates successful in the Internvl series model,  as a crucial component of our optimization pipeline. The alignment of our R1V2 model is significantly enhanced by our Skywork-VL reward model \cite{wang2025skywork}, which guides our iterative optimization process using MPO. Noticeably, the process yields a substantial reduction in the occurrence of repetitive chain-of-thought (CoT) and overthinking in the generated output.

MPO loss function can be typically expressed as follows:
\begin{equation}
\mathcal{L}=w_1\mathcal{L}_{\textit{preference}}+w_2\mathcal{L}_{\textit{quality}}+w_3\mathcal{L}_{\textit{generation}}.
\end{equation}
$\mathcal{L}_{\textit{preference}}$ is typically a DPO \cite{zhang2024direct} loss which can optimize the relative preference between positive and negative samples: 
\begin{equation}
    \mathcal{L}_{\textit{preference}}=-\log\sigma\left(\beta\log\frac{\pi_\theta\left(y_c\mid x\right)}{\pi_0\left(y_c\mid x\right)}-\beta\log\frac{\pi_\theta\left(y_r\mid x\right)}{\pi_0\left(y_r\mid x\right)}\right)
\end{equation}
Where $\beta$ is the KL penalty coefficient, and the prompt, positive sample, and negative response are represented by $x$, $y_c$, and $y_r$, respectively. The policy model $\pi_\theta$ is initialized from $\pi_0$.

$\mathcal{L}_{\textit{quality}}$ is a BCO \cite{BCO} loss. This loss helps the model to understand the absolute quality of individual responses.
This algorithm trains a binary classifier, where the logit serves as a reward and effectively
maps the chosen response to 1 and the rejected response to 0. The loss function is defined as:
\begin{align}
\mathcal{L}_{\textit{quality}} &= \mathcal{L}_{\textit{quality}}^{+} + \mathcal{L}_{\textit{quality}}^{-} \nonumber \\
&= - \Bigg[
\log \sigma\left( \beta \log \frac{\pi_{\theta}\left(y_{c} \mid x\right)}{\pi_{0}\left(y_{c} \mid x\right)} - \delta \right) 
+ \log \sigma\left( 
- \left( \beta \log \frac{\pi_{\theta}\left(y_{r} \mid x\right)}{\pi_{0}\left(y_{r} \mid x\right)} - \delta \right) 
\right)
\Bigg]
\end{align}
Here, $\delta$ is practically computed as the moving average of past rewards to stabilize training, which is a typically method in online RL method \cite{ahmadian2024back, shen2024improvingreinforcementlearninghuman}. 

Furthermore, the generation loss:
\begin{equation}
    \mathcal{L}_{\textit{generation}} = - \frac{\log \pi_{\theta}(y_{c} | x)}{|y_{c}|}
\end{equation}
 which is typically a Negative Log Likelihood loss (NLL), guiding the model to learn the chosen response, aiming to reduce the distribution shift between the base model's outputs and the preferred responses.

Mixed Preference Optimization strategy can integrate preference signals from the Skywork-VL reward model \cite{wang2025skywork} with hand-crafted rule-based constraints (e.g., format correctness, factual consistency, step-by-step reasoning completeness). This hybrid reward structure better aligns the model's outputs with both stylistic preferences and factual requirements across modalities.




\subsection{Reinforcement Fine-tuning}
In the reinforcement fine-tuning stage for VLMs, we primarily employ the GRPO Algorithm with a hybrid supervised signal composed of both rule-based reward and model-based reward. Additionally, we leverage the SSB mechanism to further enhance the efficiency of our reinforcement learning process.

\subsubsection{GRPO Algorithm with Hybrid Reward Signal}
For enhanced reasoning in multi-modal contexts, we adapt Group Relative Policy Optimization (GRPO) \cite{shao2024deepseekmathpushinglimitsmathematical}, a common RL algorithm originally developed for text-only LLMs. GRPO is a policy optimization algorithm designed to compute token-level advantage estimates by performing intra-group comparisons of generated responses conditioned on a specific query. For a given input instance $x$, a behavior policy $\pi_{\theta_{old}}$ samples a batch of $N$ candidate responses $\{y_i\}_{i=1}^{N}$. The advantage $\hat{A}_{i,t}$ for the $i$-th response at time step $t$ is determined by normalizing the rewards obtained across the response group, according to:
\begin{equation}
\hat{A}_{i,t} = \frac{r(x, y_i) - \text{mean}(\{r(x, y_1), ..., r(x, y_N)\})}{\text{std}(\{r(x, y_1), ..., r(x, y_N)\})}
\end{equation}
To mitigate the ``alignment tax'' \cite{lin2024mitigatingalignmenttaxrlhf} in reasoning ability, we again utilize our Skywork-VL reward \cite{wang2025skywork} model to introduce a preference reward signal $r_{\theta}$, supplementing the rule-based reward $r_{\textit{rule}}$. Furthermore, we incorporate a format reward $r_{\textit{format}}$ to align the model's output with the DeepSeek R1-style \cite{deepseekai2025deepseekr1incentivizingreasoningcapability} chat template. Thus, our hybrid reward function is defined as follows:
\begin{equation}
    r(x,y_i) = r_{\textit{rule}}(x, y_i) + r_{\theta}(x, y_i) + r_{\textit{format}}(x,y_i)
\end{equation}

The GRPO optimization objective incorporates a clipped surrogate loss term augmented with a KL-penalty to ensure stable policy updates, formulated as:

{\fontsize{7.5pt}{10pt}\selectfont
\begin{equation}
\mathcal{L}_{\text{GRPO}}(\theta) = \frac{1}{G} \sum_{i=1}^{G} \frac{1}{|y_i|} \sum_{t=1}^{|y_i|} 
\min \left[ 
\frac{\pi_\theta(y_{i,t} \mid x_i, y_{i,<t})}{\pi_{\theta_{\text{old}}}(y_{i,t} \mid x_i, y_{i,<t})} \hat{A}_{i,t},\ 
\text{clip} \left( 
\frac{\pi_\theta(y_{i,t} \mid x_i, y_{i,<t})}{\pi_{\theta_{\text{old}}}(y_{i,t} \mid x_i, y_{i,<t})},\ 1-\epsilon,\ 1+\epsilon 
\right) \hat{A}_{i,t}
\right]
\end{equation}
}

The hyperparameter $\epsilon$ dictates the tolerance for policy deviation. The clipping function serves to prevent excessively large policy updates by constraining the ratio of the current policy to the reference policy within a predefined interval. This mechanism promotes training stability and mitigates the risk of performance degradation resulting from overly aggressive updates.

\subsubsection{Addressing Vanishing Advantages via SSB}

However, the direct application of GRPO to VLMs encounters challenges due to its limited effectiveness, particularly after the utilization of MPO. As noted in recent work \cite{wang2025vl}, we also observed the detrimental ``Vanishing Advantages'' phenomenon during GRPO training, which can hinder the upper bound of the optimization. This issue arises when all responses within a query group converge towards uniform correctness or incorrectness, causing the relative advantage signals to diminish and impeding effective gradient-based policy updates.

\begin{figure}[h]
    \centering
    \includegraphics[width=1\linewidth]{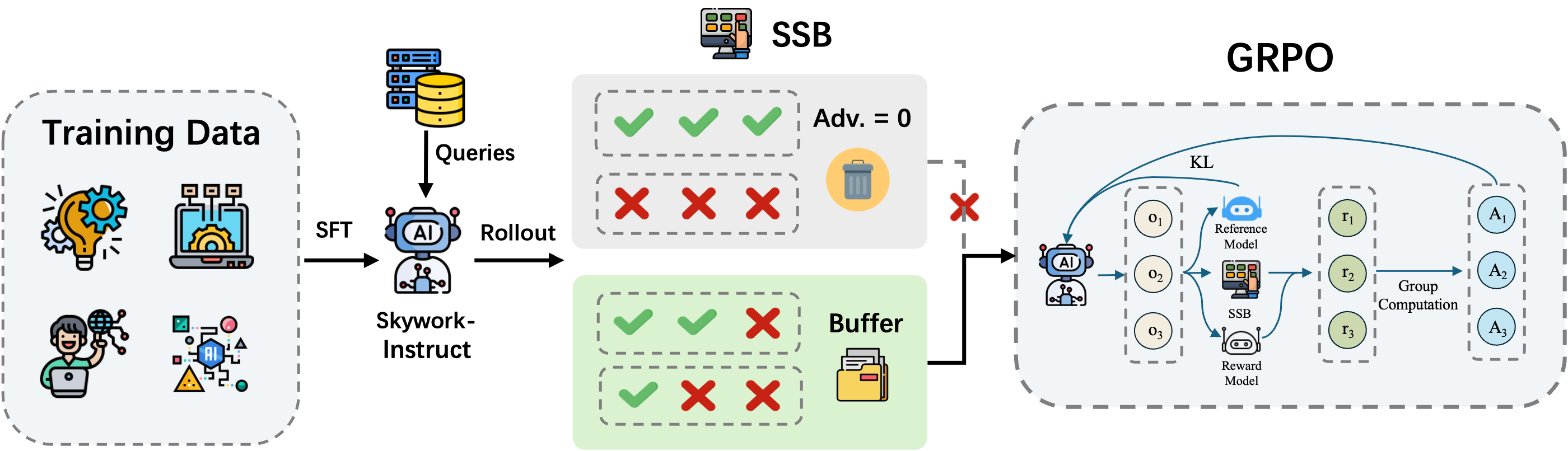}
    \caption{The Selective Sample Buffer (SSB) mechanism addresses the ``Vanishing Advantages'' problem by preserving and prioritizing high-value samples with non-zero advantages.}
    \label{fig:ssb}
\end{figure}


This approach complements concurrent efforts such as Selective Sample Replay \cite{wang2025vl}, while focusing specifically on advantage-based sample caching mechanisms. This approach prioritizes samples based on the absolute value of their advantage signals through weighted sampling, strategically reintroducing these informative samples during policy updates. This method maintains a gradient-rich training environment even as model responses converge.

A significant empirical observation is the dramatic reduction in effective training samples (those exhibiting non-zero advantages) from about 60\% at the start to under 40\% in later phases. This severe drop-off cripples training efficiency. The SSB mechanism actively combats this by guaranteeing a continuous source of valuable training signals. Moreover, we discovered that incorporating SSB into the rollout of offline inference time in advance, effectively creating a filtered prompt pool, yields a substantial improvement of over 10\% in training efficiency during the initial optimization.

Our experiments demonstrate that the SSB mechanism substantially increases training efficiency, with performance improvements equivalent to full training achieved using just a fraction of the samples. This approach not only addresses the ``Vanishing Advantages'' problem but also prevents the model from plateauing in mid-training, ensuring continuous improvement in reasoning capabilities.

Additionally, we noted that while visual reasoning and textual reasoning capabilities show complementary patterns during training, excessive emphasis on visual reasoning can lead to increased hallucination—a phenomenon we believe occurs because stronger visual reasoning necessitates more creative interpolation between visual elements. The SSB helps maintain an appropriate balance between these modalities by preserving diverse learning signals across both domains.


\section{Experiments}

We conducted comprehensive evaluations of Skywork R1V2 across multiple benchmarks designed to assess both text reasoning and visual reasoning capabilities.

\subsection{Experimental Setup}
\paragraph{Benchmarks} We evaluated R1V2 on the following benchmarks:

\subsubsection{Text Reasoning Benchmarks}
\begin{itemize}
    \item \textbf{AIME 2024} \cite{aime2024}: This benchmark includes competition problems from the 2024 American Invitational Mathematics Examination (AIME), a prestigious and highly selective contest for elite high school students. It assesses advanced mathematical competencies, requiring deep conceptual understanding and rigorous logical reasoning skills.
    
    \item \textbf{LiveCodebench} \cite{jain2024livecodebench}: A comprehensive evaluation framework for assessing coding capabilities across multiple programming languages. It features problems ranging from algorithmic challenges to software design tasks, focusing on the model's ability to understand, debug, and generate accurate code solutions under varying complexity constraints.
    
    \item \textbf{LiveBench} \cite{white2024livebench}: A dynamic reasoning benchmark designed to evaluate models' performance on everyday reasoning tasks. It combines diverse problem types including logical puzzles, causal reasoning, and counterfactual thinking to assess general reasoning capabilities beyond specialized domains.
    
    \item \textbf{IFEVAL} \cite{zhou2023instruction}: A benchmark designed to evaluate conditional reasoning in large language models, focusing on the models' ability to correctly interpret and follow ``if-then'' statements. It tests for both factual accuracy and logical consistency in responding to hypothetical scenarios.
    
    \item \textbf{BFCL} \cite{fyan2024bfcl}: The Benchmark for Faithful Chain-of-thought Language reasoning, which evaluates models on their ability to generate coherent, logically sound reasoning chains. It specifically assesses whether models can maintain logical consistency throughout multi-step reasoning processes without introducing contradictions or non-sequiturs.
\end{itemize}

\subsubsection{Multimodal Reasoning Benchmarks}
\begin{itemize}
    \item \textbf{MMMU} \cite{yue2024mmmu}: The Massive Multi-discipline Multimodal Understanding benchmark, which evaluates models across 30 academic disciplines including STEM, humanities, and social sciences. It contains college-level problems requiring both visual interpretation and domain-specific reasoning.
    
    \item \textbf{MathVista} \cite{lu2023mathvista}: A comprehensive benchmark for evaluating mathematical reasoning in visual contexts. It contains diverse mathematical problems that require interpretation of diagrams, charts, geometric figures, and real-world images, testing both perceptual understanding and mathematical problem-solving.
    
    \item \textbf{OlympiadBench} \cite{he2024olympiadbench}: A challenging benchmark comprising problems adapted from international science and mathematics olympiads. It requires advanced reasoning in physics, chemistry, biology, and mathematics, with problems presented in multimodal formats including diagrams, graphs, and equations.
    
    \item \textbf{MathVision} \cite{wang2024measuring}: A specialized benchmark focusing on geometric reasoning and spatial understanding. It contains problems that require analyzing complex geometric configurations, understanding 3D projections, and applying mathematical principles to visual scenarios.
    
    \item \textbf{MMMU-Pro} \cite{yue2024mmmup}: An extended version of the MMMU benchmark with more challenging problems and a focus on professional-level content. It emphasizes interdisciplinary reasoning and complex visual interpretation tasks that require expertise across multiple domains.
\end{itemize}

\paragraph{Evaluation Setup}
In our evaluations, the maximum generation length is set to 64K tokens for all benchmarks. We employ a unified evaluation framework using LLM Judge (OpenAI-o4) for both text reasoning and multimodal reasoning benchmarks. This approach systematically compares the model's reasoning responses with reference answers, assessing semantic equivalence and mathematical accuracy rather than relying on exact string matching. The LLM Judge can recognize when expressions differ in form but represent logically consistent solutions, making it particularly valuable for mathematical reasoning tasks where multiple solution paths or notations may be valid. For multiple-choice questions, we use a unified prompt that requires the model to provide a boxed answer format, while for open-ended questions, we instruct the model to present a step-by-step reasoning process culminating in a clearly boxed final answer. The reported performance metric is the Pass@1 score, averaged across 5 independent runs to ensure statistical reliability and account for any variations in model outputs.

\textbf{Baselines} We conduct comprehensive evaluations against several strong proprietary models, including Claude-3.5-Sonnet (20241022) \cite{Claude2024}, OpenAI-o4-mini \cite{openai2024gpt4o}, OpenAI-o1 \cite{jaech2024openai}, Gemini 2 Flash \cite{google2024gemini2}, and Kimi k1.5 longcot \cite{team2025kimi}. Additionally, we compare our method with advanced open-source models, such as Skywork-R1V1 \cite{peng2025skywork}, InternVL3-38B \cite{zhu2025internvl3}, QvQ-Preview-72B \cite{qwen2024qvq}, Deepseek R1 \cite{deepseekai2025deepseekr1incentivizingreasoningcapability} and Qwen2.5-VL-72B-Instruct \cite{bai2025qwen2}. For InternVL3-38B and other models where we report results not directly from their original papers, we conducted independent evaluations using our unified evaluation framework to ensure fair comparison.

\subsection{Main Results}

\begin{figure*}
    \centering
    \includegraphics[width=0.8\linewidth]{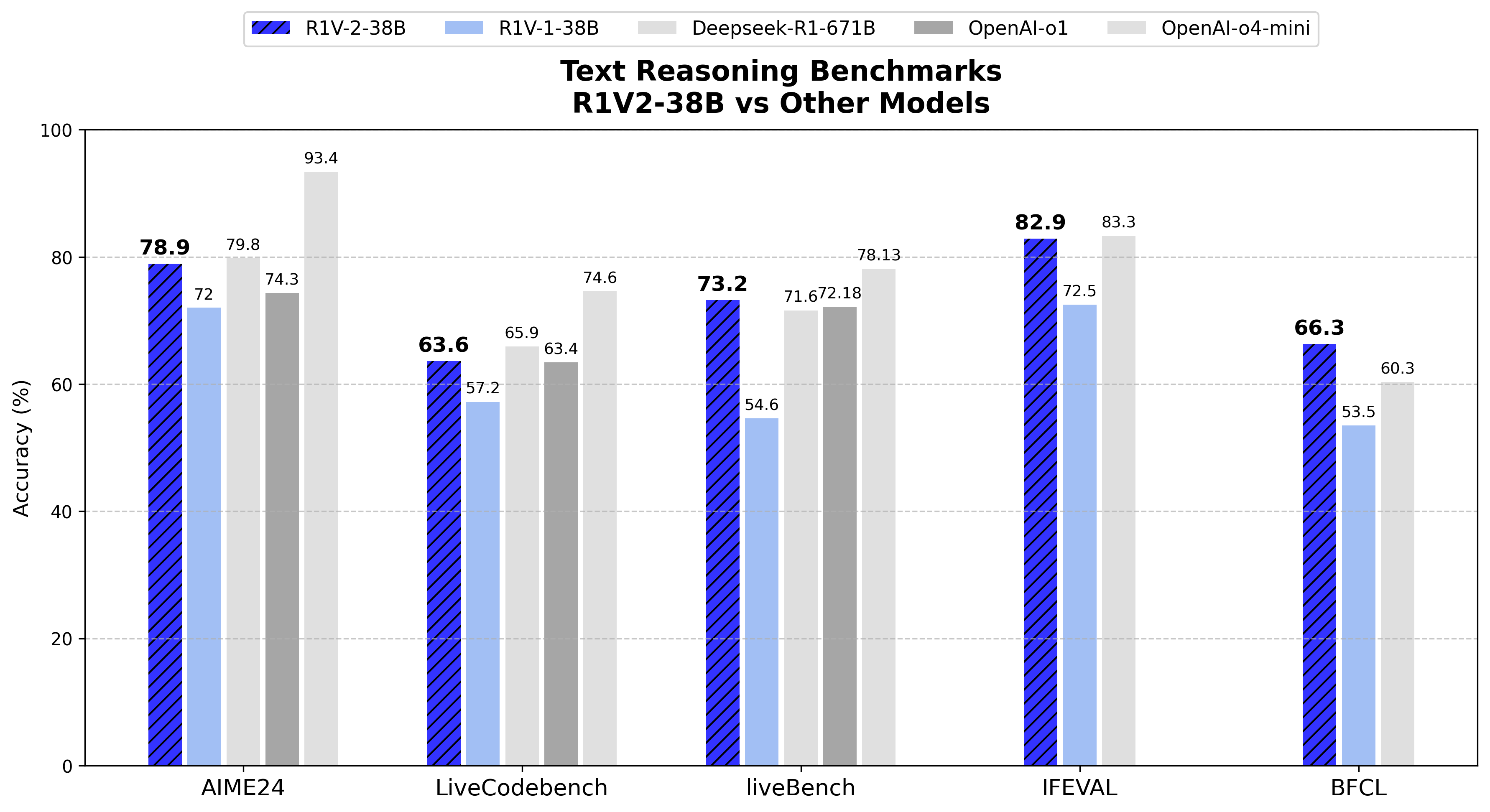}
    \caption{Performance comparison on text reasoning benchmarks.} 
    \label{fig:text_reasoning} 
\end{figure*}

\paragraph{Text Reasoning Performance}
Figure~\ref{fig:text_reasoning} presents the performance of Skywork R1V2 compared to other state-of-the-art models across various text reasoning benchmarks. Skywork R1V2 demonstrates exceptional reasoning capabilities, achieving 78.9\% on AIME24, 63.6\% on LiveCodebench, 73.2\% on LiveBench, 82.9\% on IFEVAL, and 66.3\% on BFCL.

Notably, R1V2 significantly outperforms its predecessor R1V1, with improvements of 6.9 percentage points on AIME24, 6.4 points on LiveCodebench, 18.6 points on LiveBench, 10.4 points on IFEVAL, and 12.8 points on BFCL. This consistent improvement across all benchmarks demonstrates the effectiveness of our proposed hybrid reinforcement learning approach.

When compared to larger models like Deepseek R1 (671B parameters), our R1V2 (38B parameters) achieves competitive results, even outperforming it on LiveBench (73.2\% vs. 71.6\%) and BFCL (66.3\% vs. 60.3\%). This suggests that our approach enables efficient learning with fewer parameters, making R1V2 a more practical choice for deployment scenarios with limited computational resources.

\begin{figure*}
    \centering
    \includegraphics[width=0.8\linewidth]{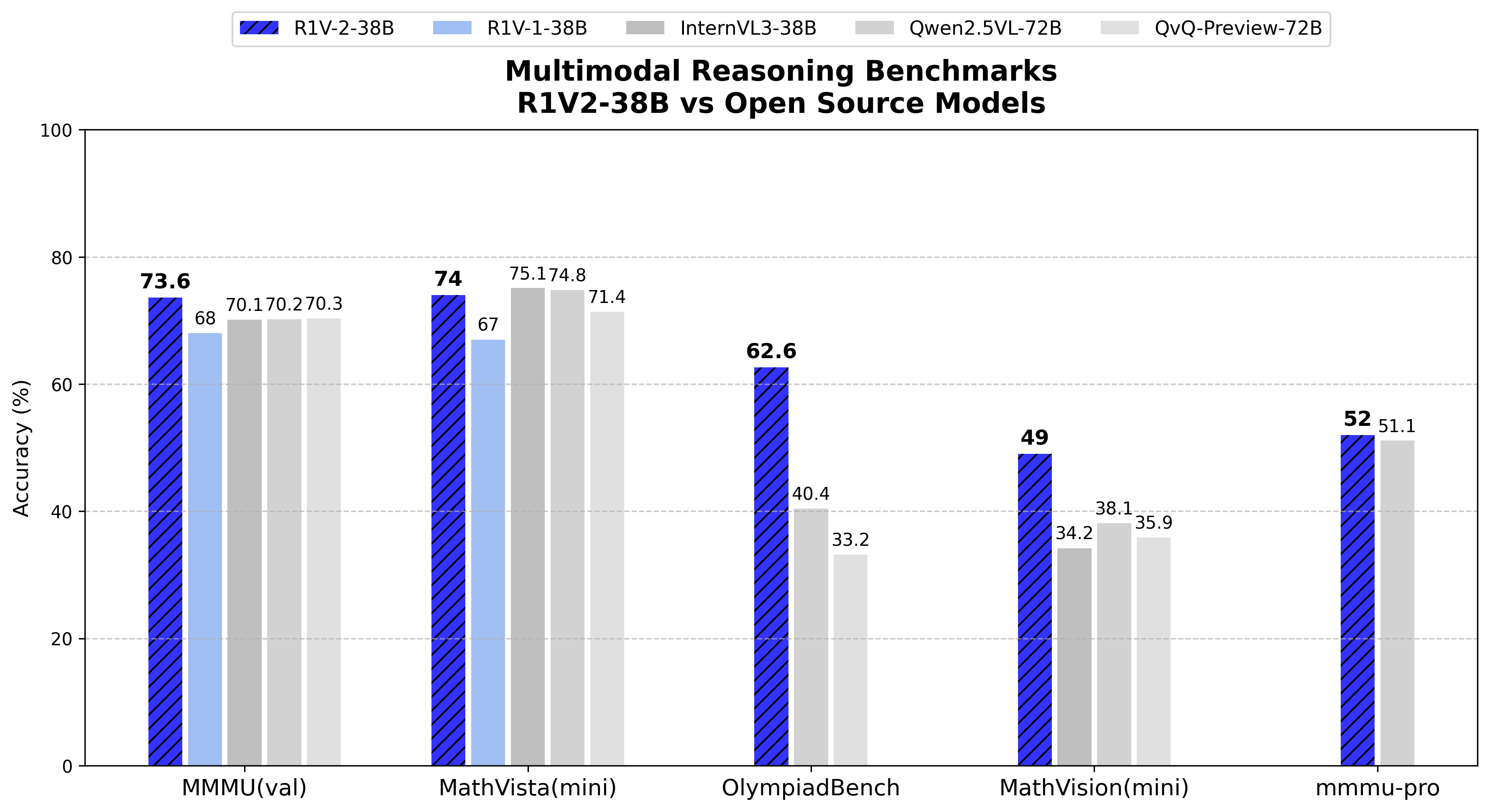}
    \caption{Comparison with open-source models on multimodal reasoning benchmarks.} 
    \label{fig:multi_reasoning_osm} 
\end{figure*}

\begin{figure*}
    \centering
    \includegraphics[width=0.8\linewidth]{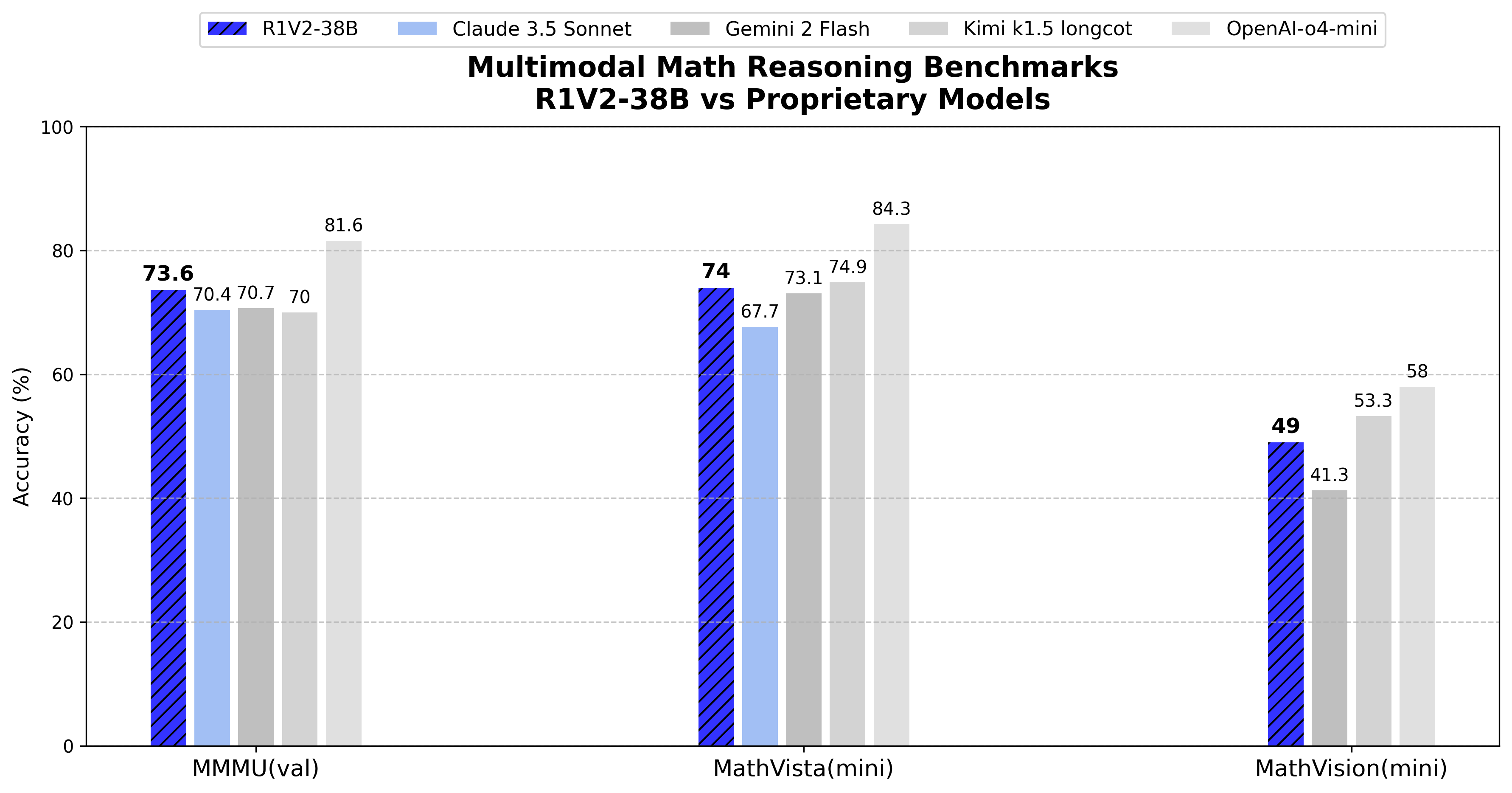}
    \caption{Comparison with proprietary models on multimodal reasoning benchmarks.} 
    \label{fig:multi_reasoning_pm} 
\end{figure*}

\paragraph{Multimodal Reasoning Performance} Figure~\ref{fig:multi_reasoning_osm} shows R1V2's performance on multimodal reasoning benchmarks compared to other open-source models. Our model achieves state-of-the-art results among open-source models of similar parameter scale, with scores of 73.6\% on MMMU, 74.0\% on MathVista, 62.6\% on OlympiadBench, 49.0\% on MathVision, and 52.0\% on MMMU-Pro.

More impressively, R1V2 outperforms even larger models such as Qwen2.5-VL-72B (70.2\% vs. 73.6\% on MMMU) and QvQ-Preview-72B (70.3\% vs. 73.6\% on MMMU).

Particularly noteworthy is R1V2's exceptional performance on OlympiadBench, where it achieves 62.6\%, substantially outperforming larger models like Qwen2.5-VL-72B (40.4\%) and QvQ-Preview-72B (33.2\%). This demonstrates R1V2's superior capability in complex mathematical reasoning tasks that require deep analytical thinking and structured problem-solving.

Figure~\ref{fig:multi_reasoning_pm} presents a comparison between R1V2 and leading proprietary models. Despite having significantly fewer parameters, R1V2 demonstrates competitive or superior performance against several commercial models.

On the MMMU benchmark, R1V2 achieves 73.6\%, outperforming Claude 3.5 Sonnet (70.4\%), Gemini 2 Flash (70.7\%), and Kimi k1.5 longcot (70.0\%). Similarly, on MathVista, R1V2's score of 74.0\% surpasses Claude 3.5 Sonnet (67.7\%) and is competitive with Gemini 2 Flash (73.1\%) and Kimi k1.5 longcot (74.9\%).

While larger proprietary models like OpenAI-o4-mini still maintain an advantage on these benchmarks, the gap has significantly narrowed compared to previous open-source efforts. This demonstrates that R1V2's innovations in reinforcement learning and multimodal integration are effectively closing the performance gap between open-source and proprietary systems.

\begin{figure*}
    \centering
    \includegraphics[width=1\linewidth]{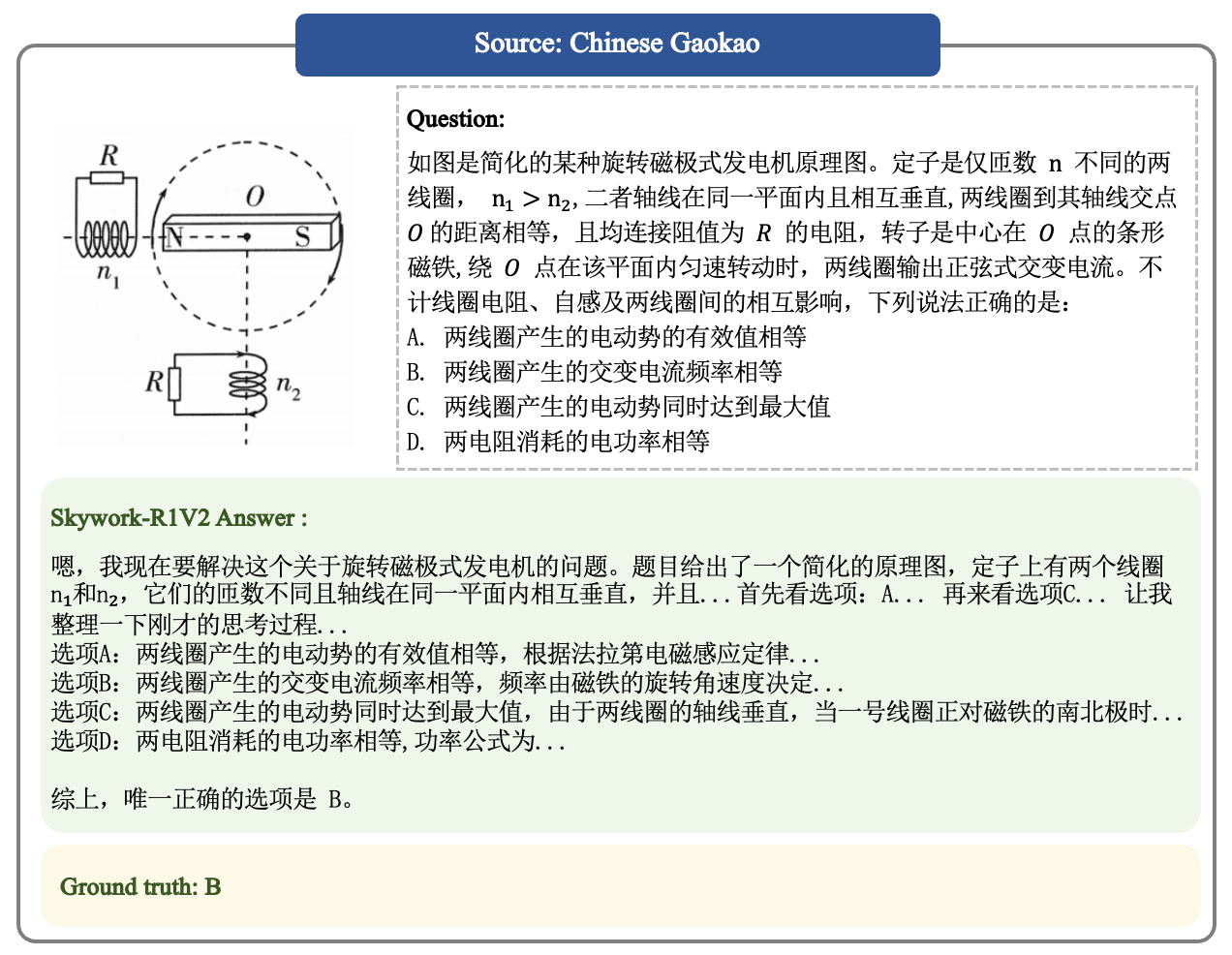}
    \caption{Reasoning Capability of Skywork-R1V2 on Chinese Gaokao Physical Problems.} 
    \label{fig:case1} 
\end{figure*}

\begin{figure*}
    \centering
    \includegraphics[width=1\linewidth]{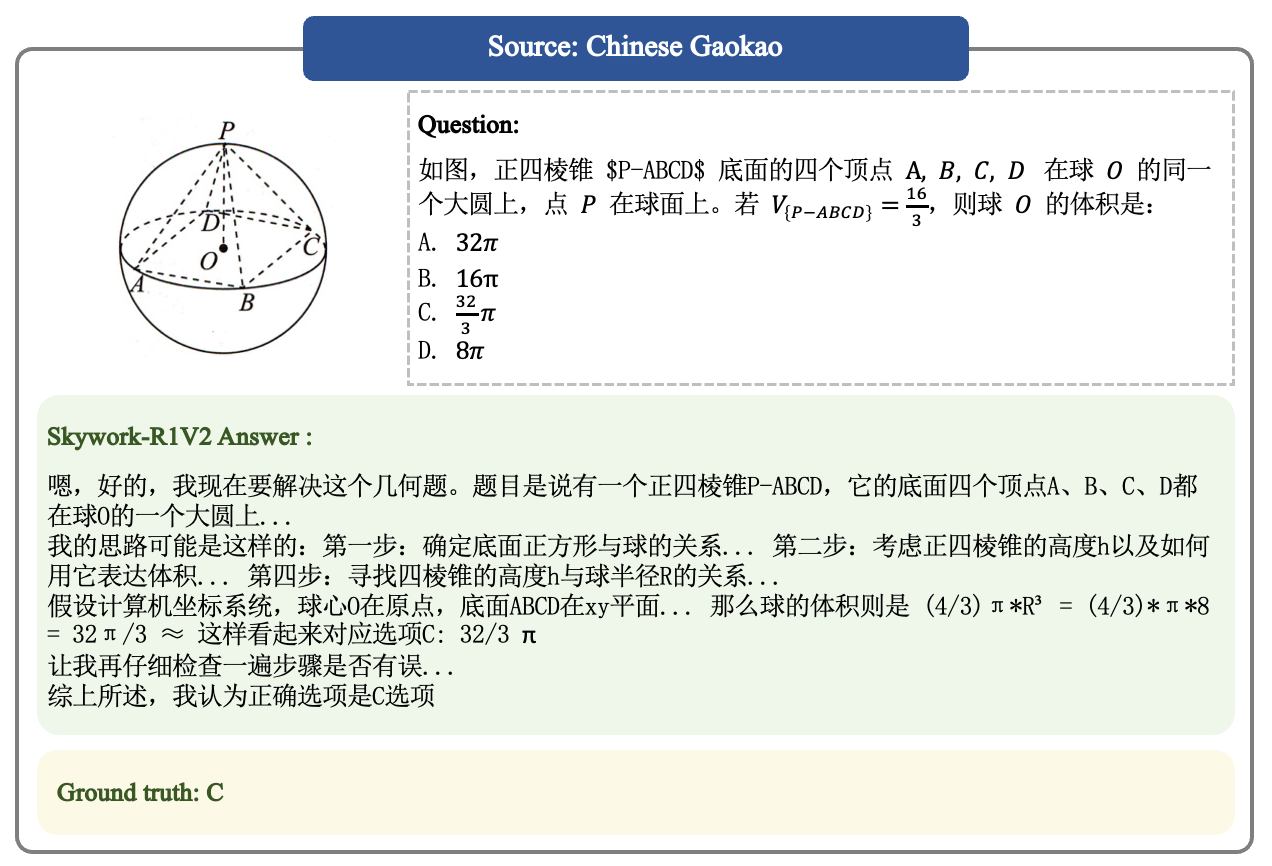}
    \caption{Reasoning Capability of Skywork-R1V2 on Chinese Gaokao Mathematical Problems.} 
    \label{fig:case2} 
\end{figure*}

\paragraph{Comprehensive Performance Analysis}
Table~\ref{tab:comprehensive_results} provides a holistic view of R1V2's performance across both text and multimodal reasoning benchmarks compared to leading open-source and proprietary models. This comprehensive analysis underscores R1V2's position as a leading open-source reasoning model with capabilities approaching or exceeding those of much larger commercial systems.

The results clearly demonstrate that R1V2 achieves state-of-the-art performance among open-source models across most benchmarks while requiring significantly fewer parameters than competitors. For example, despite being 2.9x smaller than Qwen2.5-VL-72B and QvQ-Preview-72B, R1V2 outperforms them by 3.4 and 3.3 percentage points respectively on MMMU, and by an even larger margin on OlympiadBench (62.6\% versus 40.4\% and 33.2\%).

When compared with proprietary models, R1V2 demonstrates competitive performance, particularly on mathematical and scientific reasoning tasks. Our model's performance on AIME24 (78.9\%) outperforms OpenAI-o1 (74.3\%) and approaches that of specialized proprietary systems like DeepseekR1-671B (79.8\%), confirming the effectiveness of our hybrid optimization approach in enhancing complex reasoning capabilities.

This comprehensive evaluation highlights the dual strengths of R1V2: exceptional reasoning capabilities and competitive multimodal understanding, establishing a new baseline for efficient open-source multimodal reasoning models.

\begin{table}
\centering
\small 
\setlength{\tabcolsep}{3pt} 
\caption{Comprehensive performance comparison across text and multimodal reasoning benchmarks.}
\label{tab:comprehensive_results}
\begin{tabular}{lcccccccc}
\toprule
Model & MMMU & Math- & Math- & Olympiad & AIME & LiveCode & Live & IFEVAL \\
 & & Vista & Vision & Bench & 24 & bench & Bench & \\
\midrule
\multicolumn{9}{c}{\textit{Proprietary Models}} \\
Claude-3.5-Sonnet & 70.4 & 67.7 & - & - & - & - & - & - \\
Gemini-2-Flash & 70.7 & 73.1 & 41.3 & - & - & - & - & - \\
Kimi-k1.5-longcot & 70.0 & 74.9 & 53.3 & - & - & - & - & - \\
OpenAI-o1 & - & - & - & - & 74.3 & 63.4 & 72.2 & - \\
OpenAI-o4-mini & \textbf{81.6} & \textbf{84.3} & \textbf{58.0} & - & \textbf{93.4} & \textbf{74.6} & \textbf{78.1} & - \\
\midrule
\multicolumn{9}{c}{\textit{Open-Source Models}} \\
Skywork-R1V1 & 68.0 & 67.0 & - & - & 72.0 & 57.2 & 54.6 & 72.5 \\
DeepseekR1-671B & - & - & - & - & \textbf{79.8} & \textbf{65.9} & 71.6 & \textbf{83.3} \\
InternVL3-38B & 70.1 & \textbf{75.1} & 34.2 & - & - & - & - & - \\
Qwen2.5-VL-72B & 70.2 & 74.8 & 38.1 & 40.4 & - & - & - & - \\
QvQ-Preview-72B & 70.3 & 71.4 & 35.9 & 33.2 & - & - & - & - \\
Skywork-R1V2 (Ours) & \textbf{73.6} & 74.0 & \textbf{49.0} & \textbf{62.6} & 78.9 & 63.6 & \textbf{73.2} & 82.9 \\
\bottomrule
\end{tabular}
\end{table}

\paragraph{Performance on General Vision Tasks}
While R1V2 is optimized for reasoning capabilities, we also evaluated its performance on general vision understanding tasks to assess potential trade-offs. In document understanding, R1V2 maintains competitive performance (81.3\% on AI2D \cite{kembhavi2016diagram}, 79.0\% on ChartQA \cite{masry2022chartqa} and TextVQA \cite{singh2019towards}) despite its reasoning focus, showing only a modest gap compared to specialized models. For video understanding, R1V2 achieves acceptable results (60.2\% on Video-MME \cite{fu2024videomme}, 61.5\% on MVBench \cite{li2024mvbenchcomprehensivemultimodalvideo}) but demonstrates stronger capabilities when temporal reasoning is required (1.92 on MMBench-Video \cite{fang2024mmbench} compared to InternVL2.5-38B's 1.82). 

Our analysis of hallucination tendencies revealed an important finding: while aggressive optimization for reasoning initially increased hallucination rates, our MPO approach effectively mitigated this issue, reducing hallucination from 18.4\% with standard SFT to 8.7\% with our full implementation. This balance allows R1V2 to maintain reasonable factual accuracy (68.9\% on RealWorldQA \cite{realworldqa}) while excelling at complex reasoning tasks. These results highlight the inherent trade-off between specialized reasoning and general visual understanding capabilities, with our current design deliberately prioritizing reasoning performance. Future versions will focus on enhancing general visual capabilities while preserving the strong reasoning foundation established in R1V2.

\subsection{Qualitative Analysis}  

Figure~\ref{fig:case1} demonstrates R1V2's physics reasoning capabilities through its analysis of an electromagnetic problem involving rotating magnetic fields and AC generators. When presented with a diagram showing two coils with different numbers of turns positioned perpendicular to each other, the model methodically evaluates each option by applying fundamental electromagnetic principles. R1V2 correctly identifies that while coil turns differ ($n_1 > n_2$), the frequency of the induced alternating current depends on the rotational speed of the magnetic core rather than the number of turns: ``Option B: The frequency of AC current produced by the two coils is equal; the frequency is determined by the rotational speed of the magnetic iron...'' This analysis demonstrates R1V2's ability to separate the relevant physical parameters (rotational frequency) from irrelevant ones (number of turns) when determining AC frequency, showcasing its understanding of electromagnetic induction principles. The model's systematic elimination of incorrect options based on physical laws rather than pattern matching highlights its robust scientific reasoning capabilities.

Figure~\ref{fig:case2} showcases Skywork R1V2's mathematical reasoning capabilities on a complex 3D geometry problem from the Chinese Gaokao. Faced with determining the volume of a sphere containing a square-based pyramid, R1V2 develops a structured solution approach: ``My thought process might be: First step: Determine the relationship between the square base and the sphere... Second step: Consider the height of the pyramid...'' The model strategically establishes a coordinate system, positioning the sphere's center at the origin and the base in the xy-plane, then accurately applies the volume formula $(4/3)\pi R^3 = (4/3) \cdot \pi \cdot 8 = 32\pi/3$. Particularly noteworthy is R1V2's self-verification behavior, where it explicitly states ``Let me check my steps again carefully to see if there are any errors...'' before confirming its answer. This combination of systematic problem decomposition, precise mathematical calculation, and solution verification demonstrates the model's sophisticated approach to mathematical reasoning tasks requiring spatial visualization.

\subsection{Ablation Studies}

\paragraph{Effect of Selective Sample Buffer (SSB)}

We conducted a comprehensive analysis of the Selective Sample Buffer (SSB) mechanism and its impact on training efficiency and model performance. Without SSB, we observed that the percentage of effective training samples (those with non-zero advantages) decreases substantially during training, limiting the gradient signals available for policy updates. By implementing SSB, we maintained a consistently high proportion of informative samples throughout the training process, leading to more stable optimization dynamics and better final performance.

As shown in Table~\ref{tab:ssb_ablation}, the SSB mechanism significantly improves model performance on key benchmarks. With our SSB approach, we achieved superior results on MMMU (73.6 vs. 73.4) while maintaining the strong performance on MathVista (74.0). More importantly, the proportion of effective samples remained above 60\% with SSB, compared to below 40\% without it. This efficiency gain demonstrates that SSB effectively addresses the ``Vanishing Advantages'' problem by ensuring a gradient-rich learning environment even as model responses converge.

\begin{table}
\centering
\caption{Ablation study on the effect of Selective Sample Buffer (SSB).}
\label{tab:ssb_ablation}
\begin{tabular}{lcccc}
\toprule
Method & Effective Samples  & MMMU & MathVista \\
\midrule
GRPO w/o SSB & $<40\%$ & 73.4 & 74.0 \\
GRPO w/ SSB & $>60\%$ & 73.6 & 74.0 \\
\bottomrule
\end{tabular}
\end{table}

\paragraph{SFT vs MPO vs. Hybrid}
We evaluated our proposed Hybrid approach \cite{gou2024mixed} against  Mixed Preference Optimization (MPO) and standard Supervised Fine-Tuning (SFT). Table~\ref{tab:optimization_comparison} presents the comparative results across multiple benchmarks and also reports hallucination rates. 

Although a specific configuration achieved peak MMMU performance (74.0), it incurred consistent slight performance degradation on other benchmarks, leading to its exclusion from the final release. The table shows that hallucination rate using MPO alone (8.7\%) was substantially lower than both the hybrid MPO+GRPO approach (9.1\%) and SFT (12.1\%), demonstrating MPO's strong standalone performance in reducing factual errors. On complex mathematical reasoning tasks, MPO achieved highest results on AIME 2024 (79.0) and competitive performance on OlympiadBench (60.6), significantly outperforming SFT (70.0 and 54.6, respectively). When combined with GRPO, our hybrid approach further improves generalization, achieving the highest scores on OlympiadBench (62.6) and near-optimal performance on AIME 2024(78.9). This confirms that hybrid approach successfully balances reasoning capability with generalization, addressing a key challenge faced by prior methods. Notably, SFT degraded model reasoning capabilities (as evidenced by its poor mathematical task performance), which motivated its complete removal from our subsequent training phases.

\begin{table}
\centering
\caption{Comparison of different optimization strategies.}
\label{tab:optimization_comparison}
\begin{tabular}{lccccc}
\toprule
Method & Hallucination & MMMU & MathVista & Olympiad & AIME 2024 \\
 & Rate &  &  & Bench & \\
\midrule
SFT & 12.1\% & 71.6 & 72.5 & 54.6 & 70.0 \\
MPO & \textbf{8.7\%} & 73.2 & 73.5 & 60.6 & \textbf{79.0} \\
MPO+GRPO (Ours) & 9.1\% & \textbf{73.6} & \textbf{74.0} & \textbf{62.6} & 78.9 \\

\bottomrule
\end{tabular}
\end{table}

\paragraph{Component Activation Analysis}
We examined different component activation configurations to understand their relative contributions to model performance. Our experiments varied which components were trainable during the optimization process: adapter-only, language model (LLM) + adapter, or adapter + vision encoder. The results, presented in Table~\ref{tab:component_activation}, reveal several interesting insights about the architecture's dynamics.

Adapter-only training delivered the best results across all benchmarks (73.6 on MMMU, 74.0 on MathVista, and 62.6 on OlympiadBench), while LLM + adapter and adapter + vision encoder configurations yielded lower performance. Somewhat surprisingly, activating the vision encoder during training provided minimal additional benefits compared to adapter-only training, suggesting that the primary gains come from improving the alignment between visual features and language processing rather than enhancing the visual encoding itself. These findings indicate that capabilities in text and vision are highly transferable—training one modality can directly benefit the other—and provide valuable guidance for future architectural optimizations.

\begin{table}
\centering
\caption{Performance with different component activation configurations.}
\label{tab:component_activation}
\begin{tabular}{lccc}
\toprule
Configuration & MMMU & MathVista & OlympiadBench \\
\midrule
LLM + Adapter & 72.1 & 72.0 & 60.8 \\
Adapter + Vision encoder & 72.3 & 71.8 & 60.2 \\
Adapter only & \textbf{73.6} & \textbf{74.0} & \textbf{62.6} \\
\bottomrule
\end{tabular}
\end{table}

\paragraph{MPO Threshold Analysis}
We analyzed the impact of different MPO threshold values on model performance across training iterations. With a threshold of 15, we observed more stable training dynamics compared to lower thresholds. As shown in our experiments, the model achieved 73.3\% on MathVista and 72.6\% on MMMU after 420 iterations, compared to 70.4\% and 67.8\% respectively with a threshold of 7 at iteration 2420. This indicates that higher thresholds lead to more selective and effective preference learning, resulting in better final performance with fewer iterations.

The threshold analysis revealed an interesting pattern of initial performance improvement followed by degradation with lower thresholds. With threshold 10, performance on MMMU peaked at iteration 750 with 73.6\% accuracy but declined to 68.9\% by iteration 2420. In contrast, higher thresholds maintained more consistent performance throughout training. This phenomenon aligns with our observation regarding reward hacking in iterative DPO, where excessive reinforcement can lead to overfitting to reward signals at the expense of generalization capability.

\section{Conclusion and Future Work}

We have presented Skywork R1V2, a next-generation multimodal reasoning model that addresses the fundamental challenge of balancing specialized reasoning capabilities with broad generalization. Through our novel hybrid reinforcement learning approach combining GRPO, SSB, and MPO, R1V2 achieves significant improvements across multiple reasoning and visual benchmarks, setting new standards for open-source multimodal models.

Mixed Preference Optimization (MPO) strengthens the R1V model’s alignment via the Skywork-VL reward model \cite{wang2025skywork}, effectively reducing repetitive reasoning while maintaining robust generalization. To address the challenge of ``Vanishing Advantages'' in GRPO training, we introduce the Selective Sample Buffer (SSB), which strategically retains high-quality examples with distinct advantage signals, enabling stable policy updates via gradient-rich training samples.

Our findings highlight an important trade-off between reasoning capability and visual hallucination, underscoring the need for careful reward calibration during reinforcement learning. This observation provides valuable guidance for future research in multimodal model development.

Skywork R1V2 establishes new open-source baselines with performance scores of 62.6\% on OlympiadBench, 78.9\% on AIME2024, 63.6\% on LiveCodeBench, and 73.6\% on MMMU. These results not only outperform existing open-source models but also substantially reduce the gap with proprietary state-of-the-art systems.

In future work, we plan to explore more sophisticated integration mechanisms between visual and textual modalities, further refine the balance between reasoning and generalization, and extend our hybrid reinforcement learning approach to additional domains and modalities.

\bibliography{main}

\begin{thebibliography}{10}

\bibitem{ahmadian2024back}
Arash Ahmadian, Chris Cremer, Matthias Gall{\'e}, Marzieh Fadaee, Julia Kreutzer, Olivier Pietquin, Ahmet {\"U}st{\"u}n, and Sara Hooker.
\newblock Back to basics: Revisiting reinforce style optimization for learning from human feedback in llms.
\newblock {\em arXiv preprint arXiv:2402.14740}, 2024.

\bibitem{Claude2024}
Anthropic.
\newblock Claude-3.5, 2024.

\bibitem{Qwen-VL}
Jinze Bai, Shuai Bai, Shusheng Yang, Shijie Wang, Sinan Tan, Peng Wang, Junyang Lin, Chang Zhou, and Jingren Zhou.
\newblock Qwen-vl: A versatile vision-language model for understanding, localization, text reading, and beyond.
\newblock {\em arXiv preprint arXiv:2308.12966}, 2023.

\bibitem{bai2025qwen2}
Shuai Bai, Keqin Chen, Xuejing Liu, Jialin Wang, Wenbin Ge, Sibo Song, Kai Dang, Peng Wang, Shijie Wang, Jun Tang, et~al.
\newblock Qwen2. 5-vl technical report.
\newblock {\em arXiv preprint arXiv:2502.13923}, 2025.

\bibitem{SFTorRL}
Hardy Chen, Haoqin Tu, Fali Wang, Hui Liu, Xianfeng Tang, Xinya Du, Yuyin Zhou, and Cihang Xie.
\newblock Sft or rl? an early investigation into training r1-like reasoning large vision-language models, 2025.

\bibitem{chen2023internvl}
Zhe Chen, Jiannan Wu, Wenhai Wang, Weijie Su, Guo Chen, Sen Xing, Zhong Muyan, Qinglong Zhang, Xizhou Zhu, Lewei Lu, et~al.
\newblock Internvl: Scaling up vision foundation models and aligning for generic visual-linguistic tasks.
\newblock {\em arXiv preprint arXiv:2312.14238}, 2023.

\bibitem{christiano2017deep}
Paul~F Christiano, Jan Leike, Tom Brown, Miljan Martic, Shane Legg, and Dario Amodei.
\newblock Deep reinforcement learning from human preferences.
\newblock {\em Advances in neural information processing systems}, 30, 2017.

\bibitem{gemini2.5}
Google DeepMind.
\newblock Gemini 2.5: Our most intelligent ai model, 2024.

\bibitem{deepseekai2025deepseekr1incentivizingreasoningcapability}
DeepSeek-AI.
\newblock Deepseek-r1: Incentivizing reasoning capability in llms via reinforcement learning, 2025.

\bibitem{fang2024mmbench}
Xinyu Fang, Kangrui Mao, Haodong Duan, Xiangyu Zhao, Yining Li, Dahua Lin, and Kai Chen.
\newblock Mmbench-video: A long-form multi-shot benchmark for holistic video understanding.
\newblock {\em Advances in Neural Information Processing Systems}, 37:89098--89124, 2024.

\bibitem{fu2024videomme}
Chaoyou Fu, Yuhan Dai, Yondong Luo, Lei Li, Shuhuai Ren, Renrui Zhang, Zihan Wang, Chenyu Zhou, Yunhang Shen, Mengdan Zhang, et~al.
\newblock Video-mme: The first-ever comprehensive evaluation benchmark of multi-modal llms in video analysis.
\newblock {\em arXiv preprint arXiv:2405.21075}, 2024.

\bibitem{gou2024mixed}
Qi~Gou and Cam-Tu Nguyen.
\newblock Mixed preference optimization: Reinforcement learning with data selection and better reference model.
\newblock {\em arXiv preprint arXiv:2403.19443}, 2024.

\bibitem{he2024olympiadbench}
Chaoqun He, Renjie Luo, Yuzhuo Bai, Shengding Hu, Zhen~Leng Thai, Junhao Shen, Jinyi Hu, Xu~Han, Yujie Huang, Yuxiang Zhang, et~al.
\newblock Olympiadbench: A challenging benchmark for promoting agi with olympiad-level bilingual multimodal scientific problems.
\newblock {\em arXiv preprint arXiv:2402.14008}, 2024.

\bibitem{jaech2024openai}
Aaron Jaech, Adam Kalai, Adam Lerer, Adam Richardson, Ahmed El-Kishky, Aiden Low, Alec Helyar, Aleksander Madry, Alex Beutel, Alex Carney, et~al.
\newblock Openai o1 system card.
\newblock {\em arXiv preprint arXiv:2412.16720}, 2024.

\bibitem{jain2024livecodebench}
Naman Jain, King Han, Alex Gu, Wen-Ding Li, Fanjia Yan, Tianjun Zhang, Sida Wang, Armando Solar-Lezama, Koushik Sen, and Ion Stoica.
\newblock Livecodebench: Holistic and contamination free evaluation of large language models for code.
\newblock {\em arXiv preprint arXiv:2403.07974}, 2024.

\bibitem{aime2024}
Minghui Jia.
\newblock Aime 2024 dataset.
\newblock \url{https://huggingface.co/datasets/Maxwell-Jia/AIME_2024}, 2025.

\bibitem{BCO}
Seungjae Jung, Gunsoo Han, Daniel~Wontae Nam, and Kyoung-Woon On.
\newblock Binary classifier optimization for large language model alignment, 2024.

\bibitem{kembhavi2016diagram}
Aniruddha Kembhavi, Mike Salvato, Eric Kolve, Minjoon Seo, Hannaneh Hajishirzi, and Ali Farhadi.
\newblock A diagram is worth a dozen images.
\newblock In {\em Computer Vision--ECCV 2016: 14th European Conference, Amsterdam, The Netherlands, October 11--14, 2016, Proceedings, Part IV 14}, pages 235--251. Springer, 2016.

\bibitem{li2024mvbenchcomprehensivemultimodalvideo}
Kunchang Li, Yali Wang, Yinan He, Yizhuo Li, Yi~Wang, Yi~Liu, Zun Wang, Jilan Xu, Guo Chen, Ping Luo, Limin Wang, and Yu~Qiao.
\newblock Mvbench: A comprehensive multi-modal video understanding benchmark, 2024.

\bibitem{lin2024mitigatingalignmenttaxrlhf}
Yong Lin, Hangyu Lin, Wei Xiong, Shizhe Diao, Jianmeng Liu, Jipeng Zhang, Rui Pan, Haoxiang Wang, Wenbin Hu, Hanning Zhang, Hanze Dong, Renjie Pi, Han Zhao, Nan Jiang, Heng Ji, Yuan Yao, and Tong Zhang.
\newblock Mitigating the alignment tax of rlhf, 2024.

\bibitem{liu2023visual}
Haotian Liu, Chunyuan Li, Qingyang Wu, and Yong~Jae Lee.
\newblock Visual instruction tuning.
\newblock {\em Advances in neural information processing systems}, 36:34892--34916, 2023.

\bibitem{lu2023mathvista}
Pan Lu, Hritik Bansal, Tony Xia, Jiacheng Liu, Chunyuan Li, Hannaneh Hajishirzi, Hao Cheng, Kai-Wei Chang, Michel Galley, and Jianfeng Gao.
\newblock Mathvista: Evaluating mathematical reasoning of foundation models in visual contexts.
\newblock {\em arXiv preprint arXiv:2310.02255}, 2023.

\bibitem{masry2022chartqa}
Ahmed Masry, Do~Xuan Long, Jia~Qing Tan, Shafiq Joty, and Enamul Hoque.
\newblock Chartqa: A benchmark for question answering about charts with visual and logical reasoning.
\newblock {\em arXiv preprint arXiv:2203.10244}, 2022.

\bibitem{openai2024gpt4o}
OpenAI.
\newblock Gpt-4o system card, 2024.

\bibitem{openai2025gpto4}
OpenAI.
\newblock Introducing openai o3 and o4-mini, 2024.

\bibitem{peng2025skywork}
Yi~Peng, Chris, Xiaokun Wang, Yichen Wei, Jiangbo Pei, Weijie Qiu, Ai~Jian, Yunzhuo Hao, Jiachun Pan, Tianyidan Xie, Li~Ge, et~al.
\newblock Skywork r1v: Pioneering multimodal reasoning with chain-of-thought.
\newblock {\em arXiv preprint arXiv:2504.05599}, 2025.

\bibitem{rafailov2023direct}
Rafael Rafailov, Archit Sharma, Eric Mitchell, Christopher~D Manning, Stefano Ermon, and Chelsea Finn.
\newblock Direct preference optimization: Your language model is secretly a reward model.
\newblock {\em Advances in Neural Information Processing Systems}, 36:53728--53741, 2023.

\bibitem{shao2024deepseekmathpushinglimitsmathematical}
Zhihong Shao, Peiyi Wang, Qihao Zhu, Runxin Xu, Junxiao Song, Xiao Bi, Haowei Zhang, Mingchuan Zhang, Y.~K. Li, Y.~Wu, and Daya Guo.
\newblock Deepseekmath: Pushing the limits of mathematical reasoning in open language models, 2024.

\bibitem{shen2024improvingreinforcementlearninghuman}
Wei Shen, Xiaoying Zhang, Yuanshun Yao, Rui Zheng, Hongyi Guo, and Yang Liu.
\newblock Improving reinforcement learning from human feedback using contrastive rewards, 2024.

\bibitem{singh2019towards}
Amanpreet Singh, Vivek Natarajan, Meet Shah, Yu~Jiang, Xinlei Chen, Dhruv Batra, Devi Parikh, and Marcus Rohrbach.
\newblock Towards vqa models that can read.
\newblock In {\em Proceedings of the IEEE/CVF conference on computer vision and pattern recognition}, pages 8317--8326, 2019.

\bibitem{sun2023aligning}
Zhiqing Sun, Sheng Shen, Shengcao Cao, Haotian Liu, Chunyuan Li, Yikang Shen, Chuang Gan, Liang-Yan Gui, Yu-Xiong Wang, Yiming Yang, et~al.
\newblock Aligning large multimodal models with factually augmented rlhf.
\newblock {\em arXiv preprint arXiv:2309.14525}, 2023.

\bibitem{google2024gemini2}
Gemini Team.
\newblock Introducing gemini 2.0: our new ai model for the agentic era.
\newblock \url{https://blog.google/technology/google-deepmind/google-gemini-ai-update-december-2024/\#ceo-message}, 2024.

\bibitem{team2024gemini}
Gemini Team, Petko Georgiev, Ving~Ian Lei, Ryan Burnell, Libin Bai, Anmol Gulati, Garrett Tanzer, Damien Vincent, Zhufeng Pan, Shibo Wang, et~al.
\newblock Gemini 1.5: Unlocking multimodal understanding across millions of tokens of context.
\newblock {\em arXiv preprint arXiv:2403.05530}, 2024.

\bibitem{team2025kimi}
Kimi Team, Angang Du, Bofei Gao, Bowei Xing, Changjiu Jiang, Cheng Chen, Cheng Li, Chenjun Xiao, Chenzhuang Du, Chonghua Liao, et~al.
\newblock Kimi k1. 5: Scaling reinforcement learning with llms.
\newblock {\em arXiv preprint arXiv:2501.12599}, 2025.

\bibitem{qwen2024qvq}
Qwen Team.
\newblock Qvq: To see the world with wisdom.
\newblock \url{https://qwenlm.github.io/blog/qvq-72b-preview/}, 2024.

\bibitem{qwq32b}
Qwen Team.
\newblock Qwq-32b: Embracing the power of reinforcement learning, March 2025.

\bibitem{wang2025vl}
Haozhe Wang, Chao Qu, Zuming Huang, Wei Chu, Fangzhen Lin, and Wenhu Chen.
\newblock Vl-rethinker: Incentivizing self-reflection of vision-language models with reinforcement learning.
\newblock {\em arXiv preprint arXiv:2504.08837}, 2025.

\bibitem{wang2024measuring}
Ke~Wang, Junting Pan, Weikang Shi, Zimu Lu, Houxing Ren, Aojun Zhou, Mingjie Zhan, and Hongsheng Li.
\newblock Measuring multimodal mathematical reasoning with math-vision dataset.
\newblock {\em Advances in Neural Information Processing Systems}, 37:95095--95169, 2024.

\bibitem{wang2025enhancingreasoningabilitymultimodal}
Weiyun Wang, Zhe Chen, Wenhai Wang, Yue Cao, Yangzhou Liu, Zhangwei Gao, Jinguo Zhu, Xizhou Zhu, Lewei Lu, Yu~Qiao, and Jifeng Dai.
\newblock Enhancing the reasoning ability of multimodal large language models via mixed preference optimization, 2025.

\bibitem{wang2025skywork}
Xiaokun Wang, Jiangbo Pei, Wei Shen, Yi~Peng, Yunzhuo Hao, Weijie Qiu, Ai~Jian, Tianyidan Xie, Xuchen Song, Yang Liu, et~al.
\newblock Skywork-vl reward: An effective reward model for multimodal understanding and reasoning.
\newblock {\em arXiv preprint arXiv:2505.07263}, 2025.

\bibitem{white2024livebench}
Colin White, Samuel Dooley, Manley Roberts, Arka Pal, Ben Feuer, Siddhartha Jain, Ravid Shwartz-Ziv, Neel Jain, Khalid Saifullah, Siddartha Naidu, et~al.
\newblock Livebench: A challenging, contamination-free llm benchmark.
\newblock {\em arXiv preprint arXiv:2406.19314}, 2024.

\bibitem{realworldqa}
X.AI.
\newblock Grok-1.5 vision preview.
\newblock \url{https://x.ai/blog/grok-1.5v}, 2024.

\bibitem{fyan2024bfcl}
Fanjia Yan, Huanzhi Mao, Charlie Cheng-Jie Ji, Tianjun Zhang, Shishir~G. Patil, Ion Stoica, and Joseph~E. Gonzalez.
\newblock Berkeley function calling leaderboard.
\newblock \url{https://gorilla.cs.berkeley.edu/blogs/8_berkeley_function_calling_leaderboard.html}, 2024.

\bibitem{yue2024mmmu}
Xiang Yue, Yuansheng Ni, Kai Zhang, Tianyu Zheng, Ruoqi Liu, Ge~Zhang, Samuel Stevens, Dongfu Jiang, Weiming Ren, Yuxuan Sun, et~al.
\newblock Mmmu: A massive multi-discipline multimodal understanding and reasoning benchmark for expert agi.
\newblock In {\em Proceedings of the IEEE/CVF Conference on Computer Vision and Pattern Recognition}, pages 9556--9567, 2024.

\bibitem{yue2024mmmup}
Xiang Yue, Tianyu Zheng, Yuansheng Ni, Yubo Wang, Kai Zhang, Shengbang Tong, Yuxuan Sun, Botao Yu, Ge~Zhang, Huan Sun, et~al.
\newblock Mmmu-pro: A more robust multi-discipline multimodal understanding benchmark.
\newblock {\em arXiv preprint arXiv:2409.02813}, 2024.

\bibitem{zhang2024direct}
Ruohong Zhang, Liangke Gui, Zhiqing Sun, Yihao Feng, Keyang Xu, Yuanhan Zhang, Di~Fu, Chunyuan Li, Alexander Hauptmann, Yonatan Bisk, et~al.
\newblock Direct preference optimization of video large multimodal models from language model reward.
\newblock {\em arXiv preprint arXiv:2404.01258}, 2024.

\bibitem{zhou2023instruction}
Jeffrey Zhou, Tianjian Lu, Swaroop Mishra, Siddhartha Brahma, Sujoy Basu, Yi~Luan, Denny Zhou, and Le~Hou.
\newblock Instruction-following evaluation for large language models.
\newblock {\em arXiv preprint arXiv:2311.07911}, 2023.

\bibitem{zhu2025internvl3}
Jinguo Zhu, Weiyun Wang, Zhe Chen, Zhaoyang Liu, Shenglong Ye, Lixin Gu, Yuchen Duan, Hao Tian, Weijie Su, Jie Shao, et~al.
\newblock Internvl3: Exploring advanced training and test-time recipes for open-source multimodal models.
\newblock {\em arXiv preprint arXiv:2504.10479}, 2025.

\end{thebibliography}
\bibliographystyle{plain}
\end{document}